\pdfoutput=1

\documentclass[11pt]{article}

\usepackage[final]{acl}
\usepackage{tcolorbox}
\usepackage{booktabs}
\usepackage{times}
\usepackage{comment}
\usepackage{latexsym}
\usepackage{amssymb}
\usepackage{makecell}

\usepackage[T1]{fontenc}

\usepackage[utf8]{inputenc}

\usepackage{microtype}
\usepackage{amsmath} 
\usepackage{inconsolata}

\usepackage{graphicx}

\title{Fin-Bias: Comprehensive Evaluation for LLM Decision-Making under human bias in Finance Domain}

  \author{Xiaoyu Hu\thanks{~~Corresponding Author}$^1$ and Jinman Zhao$^2$\\
  $^1$Rutgers University, \\$^2$Department of Computer Science, University of Toronto \\
  \texttt{xh391@scarletmail.rutgers.edu, jzhao@cs.toronto.edu}
\\}

\begin{document}
\maketitle
\begin{abstract}
Large language models (LLMs) are increasingly deployed in financial contexts, raising critical concerns about reliability, alignment, and susceptibility to adversarial manipulation. While prior finance-related benchmarks assess LLMs' capabilities in stock trading, they are often restricted to small sample and fail to demonstrate LLM susceptibility to context with potential human bias. We introduce Fin-Bias (financial herding under long and uncertain financial context), a benchmark for evaluating LLM investment decision-making when faced with uncertainty and possible human-biased opinions. Fin-Bias includes 8868 long firm-specific analyst reports, including firm aspects summarized and analyzed by sophisticated analysts with investment ratings (Bullish/Neutral/Bearish) spanning from various industries. We present large language models with firm analyst reports with/without analyst investment ratings and even with 'fake' rating, to get investment ratings generated by LLMs. Our results reveal that LLMs tend to herd the explicit bias in context. We also develop a method to detect potential human opinions, which can encourage LLMs to think independently, some models even exceed human performance in predicting future stock return.  Our code is available at \url{https://github.com/Xiaoyu1216/Fin-Bias.git}.
\end{abstract}

\begin{table*}[htbp]
\centering

\resizebox{\textwidth}{!}{
\begin{tabular}{lcccccc}
\toprule
\textbf{Benchmark} & \textbf{Long Context} & \textbf{Summarize} &
\begin{tabular}[c]{@{}c@{}}\textbf{Human}\\ \textbf{Bias}\end{tabular} &
\begin{tabular}[c]{@{}c@{}}\textbf{Large}\\ \textbf{Sample}\end{tabular} &
\begin{tabular}[c]{@{}c@{}}\textbf{Decision-}\\ \textbf{Making}\end{tabular} \\
\midrule
FinQA~\cite{chen2021finqa} & $\times$ & $\times$ & $\times$ & $\checkmark$ & $\times$  \\
ConvFinQA~\cite{chen2022convfinqa} & $\checkmark$ & $\times$ & $\times$ & $\checkmark$ & $\times$  \\
FiQA-SA~\cite{maia2018financial} & $\times$ & $\times$ & $\checkmark$ & $\checkmark$ & $\times$  \\
FPB~\citet{malo2014good} & $\times$ & $\times$ & $\checkmark$ & $\checkmark$ & $\times$  \\
ACL18~\cite{xu2018stock} & $\checkmark$ & $\times$ & $\checkmark$ & $\times$ & $\checkmark$  \\
INVESTORBENCH~\cite{li2025investorbench} & $\checkmark$ & $\checkmark$ & $\times$ & $\times$ & $\checkmark$  \\
Fintrade~\cite{xie2024finben} & $\checkmark$ & $\checkmark$ & $\times$ & $\times$ & $\checkmark$  \\
Fin-Bias (Ours) & $\checkmark$ & $\checkmark$ & $\checkmark$ & $\checkmark$ & $\checkmark$ \\
\bottomrule
\end{tabular}}
\caption{Comparison with existing Financial Benchmarks. "Summarize" refers to the context is summarized firm events related to price. "Human bias" refers to the context contains human opinions like tweets, analyst report etc. "Large Sample" refers to the number of unique firms in sample is large.}\label{tab:long_context_benchmarks}
\end{table*}

\section{Introduction}

The advancements in large language models(LLMs) have gained significant attention for their application in various domains. Recent studies have studied the use of LLMs in the finance domain, such as financial sentiment analysis ~\citep{wang2023fingpt}, finance question answering ~\citep{zhao2024docmath}, and stock trading ~\citep{yu2024fincon,wang-etal-2024-gpt}. While the existing literature  ~\citep{chen2021finqa,chen2022convfinqa,mateega2025financeqa} has provided a broad range of evaluation benchmarks in the finance domain, few of them evaluate LLM decision-making behavior in finance domain. Even though some create stock trading tasks ~\citep{li2025investorbench,xie2024finben}, none of them evaluate LLM decision-making under context with potential human bias. Decision-making requires identifying relevant signals, reconciling conflicting evidence, and synthesizing heterogeneous data into a coherent judgment, which represents the core
intellectual skills that large language model (LLM) agents must master—information gathering, coordination,
and grounded reasoning. Some studies (e.g., develop advanced LLM agent frameworks to combine and analyze financial documents from various sources, including news, financial reports, social media etc. A major problem is that some financial documents contain individual human opinions which are likely to be biased. For instance, analysts tend to provide 'bullish' rather than 'bearish' ratings for firms due to overconfidence or strategic incentives. Whether LLMs will herd to human-biased opinions still remains unresolved. If LLMs do herd human bias a lot, LLMs will fail to give the right answer when most 'people' are wrong which is not uncommon in financial market trading. 

To bridge the gap, we introduce Fin-Bias, a financial investment-rating dataset comprising 8,868 input–label pairs. Each input is a long-form, firm-specific analyst report, produced from in-depth research, covering firms across multiple sectors, including Technology, Industrials, Financial Services, Healthcare, Utilities, Energy, Communication Services, and Real Estate. The first sentence of each report explicitly states the analyst’s investment rating for the covered firm (Bullish / Neutral / Bearish). We use Fin-Bias to study LLM herding behavior, the tendency of models to follow human opinions present in context rather than produce independent judgments. To demonstrate the exact influence of human opinions on LLM decision-making, we create two perturbations of each report: (1) removal of the first sentence that contains the analyst rating; and (2) retention of the first sentence but replacement of the original rating with a 'fake' rating different from the original one. Compared to existing financial benchmarks (see Table~\ref{tab:long_context_benchmarks}) and in line with recent calls for more comprehensive benchmark design~\citep{peng-etal-2025-m3gqa}, Fin-Bias consists of a large sample of long-context inputs which are comprehensively summarized firm aspects related to stock price, unlike short sentences that contain limited evidence but only human sentiments/opinions from tweets. For evaluation, we prompt LLMs with a chain-of-thought instruction and feed them three variants of each report (original, analyst rating removed, and analyst rating replaced with a 'fake' rating). We then compare the LLM decision-making across these conditions to quantify the influence of explicit human bias on model decisions. We introduce \textbf{Herding Score}, which captures how much LLM ratings align with human (analyst ratings or 'fake' ratings). The results suggest to some extent LLMs tend to herd human opinions in decision-making, the herding score increases a lot when including analyst/'fake' ratings in context. The difference ranges from 5\% to 10\% depending on models. Besides, we use a fine-grained quantile long-short portfolio method to get ground-truth investment rating label based on real stock market return to make comparison between LLM performance and human performance in stock trading. The results indicate that when analyst rating is present in context, all models have nearly the same performance as the analyst, achieving around around 33\% accuracy in providing investment rating. We also find when the analyst rating is excluded in context, some models' performance exceeds analysts, while some models' performance drop. The model herding does not depend on model size, even cutting-edge complex models such as GPT-5, GPT-4 have possibilities to herd fake rating. To alleviate model herding, we propose to filter sentences that contain strong opinions out of the context based on Multi-Perspective Question Answering (MPQA) Subjectivity Lexicon. The results indicate the method can help improve model performance by 2-4 points especially for light open-source model like Qwen3-8B.   
\\

The main contribution of our work is three-fold: 
\begin{itemize}
    \item We conduct a comprehensive study of 18 models (Open Source \& Private), showing that models are likely to herd human opinions even when human opinions take very small part in the context. 
    \item We propose the Fin-Bias benchmark dataset, which contains long-form, summarized, time-sensitive and potentially biased firm analysis across different industries to evaluate LLM financial reasoning capacities.
    \item We propose a potential way to instruct large language models to avoid herding human opinions and improve independent thinking and evaluate the LLM performance using human performance as benchmark rather than base model.
\end{itemize}

\section{Related Work}

\subsection{Financial Evaluation Benchmarks}

Financial texts is one of the most common unstructured data formats in the finance domain. There is a large strand of literature studying how to make use of large language models in performing financial textual analysis. One of major tasks is financial sentiment analysis, which aims to detect sentiment (positive, negative, or neutral) from financial texts and is used for investment decision-making. Prior studies provide limited financial analysis benchmark datasets, including news headline classification ~\citep{alvarado2015domain}, named entity recognition (NER) ~\citep{alvarado2015domain}, question answering (QA) ~\citep{chen2022convfinqa}, sentiment analysis ~\citep{malo2014good, maia2018financial, cortis2017semeval}. For stock prediction, some studies use LLMs to analyze stock news and predict stock prices ~\citep{yu2025finmem,xie2024finben}. The major limitations of those datasets are as follows: first, most datasets are short-contexts-based, formed by simply concatenating a bunch of news/stock tweets. In finance domain, investors get information from a great deal of financial documents, forming a long context. The short sentences lacking specific contexts are of least helpful for evaluating large language models' capacities in assisting with making investment decisions; Even concatenating stock news forms long context, they lack summarization and deep research, making LLMs susceptible to noise. Second, most prior stock prediction benchmarks ~\citep{li2025investorbench} only select less than 20 stocks that are well-known and concentrated in the technology industry (e.g. Apple, Google, Tesla etc), failing to evaluate LLM stock analysis comprehensively. Third, financial corpus contain human bias (over-optimism/over-pessimism), however, previous benchmarks have not uncovered LLMs' capacities in correcting human bias in the finance domain. 

\subsection{LLM Strategic Decision-Making}

Strategic decision-making ~\citep{punt2017strategic} evaluates the model’s proficiency in synthesizing diverse information to formulate and implement trading strategies, a challenge even for experts. A bunch of studies provides evaluation on LLMs' decision-making behavior under uncertainty. For instance, LLMs generally exhibit patterns similar to humans, such as risk aversion and loss aversion, with a tendency to overweight small probabilities ~\citep{jia2024decision}. ~\citet{liu2024dellma} evaluate how LLMs perform in a financial investing situation with a variety of uncertainties, however, they only use historical prices as the context. In contrast, we use long plain financial texts to form context. This matters because real-world decisions often rely on long~\citep{luo2025monointernvl15cheaperfastermonolithic,li2026memorybarrierhighlymemory} and heterogeneous documents. Prior literature ~\citep{lyons2021overconfidence,bosquet2015analysts} suggest that human bias (e.g. overconfidence) may be a crucial factor for explaining how false and low-quality information spreads via news, social media, firm report etc. However, most LLM agent for trading frames ~\citep{xiao2024tradingagents} rely on those biased texts without comprehensively evaluating to what extent LLMs can correct human bias hidden in the contexts. Prior work has also shown that biases can persist in downstream language models~\citep{dai-etal-2024-mitigate}. ~\citet{huang2019reducing} suggests the existence of sentiment bias which can pose a concern for using the text generated by language models in downstream applications.

\section{Dataset Construction}

\paragraph{Data Collection.} We collect analyst reports with investment ratings sourced from Yahoo Finance. We randomly download reports from different sectors, including Technology, Industrials, Financial Services, Healthcare, Utilities, Energy, Communication Services and Real Estate. The reports are in pdf format and we extract the content from pdf using large language models. The analyst reports summarize firm comprehensive and significant aspects related to stock price and provide an investment rating(bullish/neutral/bearish) in the first sentence. The total number of analyst reports is 8868. Table \ref{tab:Sample} provides detailed statistics for each sector. Unlike prior benchmark paper, we include firms from a broad range of industries. Each report contains an average of about 4,000 tokens, providing a relatively long context for large language models. An example analyst report is shown in appendix.

\paragraph{Data Perturbation.} We evaluate the susceptibility of LLMs to human-like biases, specifically herding behavior, when making investment decisions under long-context settings by systematically perturbing the analyst reports provided to the models. In particular, we prompt LLMs to generate investment ratings based on analyst analyses that have been modified to exhibit varying degrees of bias, allowing us to assess how sensitive the models are to biased or misleading contextual cues embedded in long context. There is a large strand of literature  ~\citep{chan2007analysts,grinblatt2023analyst,bosquet2015analysts} indicating analysts forecasts are systematically biased upward due to overconfidence and strategic incentives (ties to investment banking). Table \ref{tab:Sample Analysts} suggests most analyst reports provide 'bullish' rating in our sample (72.28\%), while very few analysts provide 'bearish' rating (0.29\%), which proves the fact documented in prior literature that analysts tend to over-optimistic. To gauge different LLMs' herding behavior, we conduct two comparison experiments: first, we provide LLMs with analyst reports with/without obvious analyst ratings and compare the differences in LLM investment ratings. Although an analyst report may contain implicit biases throughout the narrative, the analyst’s investment rating (e.g., buy/bullish, hold/neutral, sell/bearish) is typically stated explicitly in the opening sentence. Therefore, in our first experiment, we perturb the original report by removing only this initial sentence that conveys the explicit investment recommendation. An example is provided in the appendix. This allows us to examine the extent to which the LLM relies on explicit human-provided views when forming its own investment decision; Second, to further evaluate LLM herding behavior, we introduce a 'fake' analyst rating by deliberately altering the rating stated in the first sentence of the original report. The 'fake' rating is arbitrary, which enables us to assess the extent to which LLMs adopt, or “herd toward” a explicit and false rating even when it contradicts the subsequent analysis. An example is provided in the appendix. 

\paragraph{Ground-truth labels.} To evaluate the practical implications of LLM herding in finance domain—specifically, whether alignment with human opinions improves or degrades investment decisions—it is necessary to obtain ground-truth investment ratings. Because analyst-issued investment ratings are known to exhibit substantial and systematic bias, they are unsuitable as ground-truth labels for empirical prediction tasks. Consequently, we rely on realized stock returns to define true outcomes. Prior studies ~\citep{liu2023fingpt, yu2024fincon} typically construct labels using daily returns combined with ad-hoc threshold rules—for example, designating returns above 1\% as “bullish,” between -1\% and 1\% as “neutral,” and below -1\% as “bearish.” However, such cutoffs are inherently arbitrary and may not accurately control for firm-specific risk and capture economically meaningful variation in firm performance. To address this limitation, we develop a fine-grained approach for constructing ground-truth investment ratings using portfolio-based alpha generation standard. The details can be found in appendix. The second half of table~\ref{tab:Sample Analysts} shows that, following analyst report date, based on our quantile-based investment rating classification method, approximately 30\% of stocks fall into the bullish category, roughly 30\% into the bearish category, and the remaining 40\% into the neutral category, indicating a relatively balanced distribution across years. More importantly, this distribution is substantially more balanced than that observed in analyst-issued ratings.

\begin{table}[t]
\centering
\caption{Sample Information}\label{tab:Sample}
\begin{tabular}{lcc}
\toprule
Sector & Num & Percentage \\
\midrule
Technology & 1491 & 16.8\% \\
Industrials & 1436 & 16.2\% \\
Financial Services & 1367 & 15.4\% \\
Healthcare & 1288& 14.5\% \\
Consumer Defensive & 982& 11.07\%\\
Utilities & 701& 7.9\%\\
Communication Services & 590& 6.6\%\\
Energy & 615& 6.9\%\\
Real Estate & 398 & 4.5\%\\
Total & 8868 & 100\% \\
\bottomrule
\end{tabular}
\end{table}

\begin{table}[ht]
\centering
\caption{Sample Investment Rating }\label{tab:Sample Analysts}
\begin{tabular}{lcc}
\toprule
Analyst Investment Rating & Num & Percentage \\
\midrule
Bullish & 6410 & 72.28\% \\
Bearish & 26 & 0.29\% \\
Neutral & 2432 & 24.74\% \\
Total & 8868 & 100\%\\
\midrule
Return-based Rating \\
\midrule
Bullish & 2664 & 30\%\\
Bearish & 2662 & 30\%\\
Neutral & 3542& 40\%\\
\bottomrule
\end{tabular}
\end{table}

\section{Experiments}

\subsection{Experiment Setup}

\textbf{Models.} We evaluate a diverse set of language models to capture a broad spectrum of capabilities, spanning both open-source and closed-source models. Specifically, we include: GPT-5.1 ~\citep{openai_gpt5}, GPT-4.1 ~\citep{openai_gpt3_4}, Claude-3.5-Haiku ~\citep{anthropic2024}, Claude-4-Sonnet ~\citep{anthropic2025}. We also include open-source models such as Meta-Llama-3-8B-Instruct~\citep{meta_llama3_8b}, Meta-Llama-3.1-8B-Instruct~\citep{meta_llama3_1_8b}, Mistral-7B-Instruct-v0.3~\citep{mistral_7b_instruct}, Mistral-Nemo-Instruct-2407~\citep{mistral_nemo_2407}, gemma-7b-instruct~\citep{gemma_7b_it}, gemma-2-9b-instruct~\citep{gemma_2_9b_it}, gemma-2-27b-instruct~\citep{gemma_2_27b_it}, Qwen2-7B-Instruct~\citep{qwen2_7b_instruct}, Qwen2.5-7B-Instruct~\citep{qwen2.5}, Qwen2.5-14B-Instruct~\citep{qwen2.5}, Qwen3-8B ~\citep{qwen3technicalreport}, DeepSeek-V2-Lite-Chat ~\citep{deepseekv2}, internlm2-chat-7b ~\citep{internlm2_chat_7b}, Yi-1.5-9B-Chat~\citep{yi_1_5_9b_chat}, glm-4-9B-chat~\citep{thudm2024glm49bchat}.
\\

\textbf{Metrics.} To capture how much LLMs herd to analysts/fake ratings when providing investment rating, we introduce \textbf{Herding Score}, which assigns 1 when a model's rating $m_i$ is as same as the rating from analyst rating/manipulated rating $a_i$ and zero otherwise and then take average. \\
\begin{equation}
\text{Herding Score} = \frac{1}{N} \sum_{i=1}^{N} I(m_i, a_i)
\end{equation}
To assess whether alignment with human opinions enhances or impairs investment decision-making, we compute the accuracy of LLM-generated investment ratings using our ground-truth labels. Higher accuracy reflects stronger model capability in predicting stock returns through financial reasoning under long, potentially biased textual inputs. The accuracy differential between predictions made with and without embedded analyst ratings captures the extent to which herding behavior influences LLM decisions. The prompt template used in this evaluation is presented in Figure~\ref{fig:prompt}.

\begin{figure}
\centering
\begin{tcolorbox}[colback=black!5!white, colframe=black, title=Chain-of-Thought Prompt]

\textbf{[System Input]:}  

"You are an investor. Analyze the firm analyst report logically. \\
Then provide your own investment rating. \\
Format as a JSON object with the following fields: \\
answer: The precise answer to the question. Only one of {bullish, neutral, bearish}. \\
reason: One or more paragraphs indicating why you provide the answer. \\

\vspace{5pt}
\textbf{[User Input]:}  

\{analyst report content\}  

\end{tcolorbox}
\caption{Prompt Template}
\label{fig:prompt}
\end{figure}

\begin{table*}[ht]
\scriptsize 
\setlength{\tabcolsep}{3pt}
\renewcommand{\arraystretch}{0.9}
\centering
\begin{tabular}{lccccc}

\toprule
\textbf{Models} & 
\textbf{Communication Services} & 
\textbf{Consumer Defensive} & \textbf{Energy} & \textbf{Financial Services} & \textbf{Healthcare} \\

\midrule
\multicolumn{6}{c}{\textit{Private Models}} \\
\midrule
\textbf{GPT-5}  & 82.7/70.2 & 93.2/88.3 & 92.3/78.9 & 95.1/87.3 & 95.3/84.1  \\
\textbf{GPT-4}  & 84.1/75.4 & 94.8/86.2 & 93.2/89.0 & 96.3/94.3 & 96.9/92.2  \\
\textbf{Claude-3.5-Haiku}  & 84.5/76.4 & 95.8/90.0 & 94.4/85.4 & 96.3/91.3 & 97.9/95.3  \\
\textbf{Claude-4-Sonnet} & 85.7/56.7 & 93.2/81.7 & 88.3/75.0 & 90.5/84.2 & 94.1/83.3   \\

\midrule
\multicolumn{6}{c}{\textit{Open-Source Models}} \\
\midrule
\textbf{gemma-7b-it} & 96.4/72.1 & 93.8/84.7 & 99.2/82.0 & 98.9/91.1 & 98.2/90.3 \\
\textbf{gemma-2-9b-it} & 95.6/68.9 & 97.6/92.2 & 96.1/77.3 & 97.9/90.9 & 97.3/90.7 \\
\textbf{gemma-2-27b-it} & 94.1/67.9 & 97.9/91.3 & 96.5/78.1 & 98.6/89.2 & 97.9/89.7 \\
\textbf{Meta-Llama-3-8B-it} & 87.2/66.3  & 96.8/86.5  & 95.0/75.8  & 96.2/89.5  & 96.4/86.9 \\
\textbf{Meta-Llama-3.1-8B-it} & 86.9/66.7 & 91.9/85.6 & 88.9/74.5 & 94.5/88.2 & 93.5/86.6  \\
\textbf{Qwen2-7B-it} & 93.5/68.6 & 94.1/88.9 & 89.9/75.7 & 96.6/86.9 & 95.1/88.1 \\
\textbf{Qwen2.5-7B-it} & 98.8/71.4 & 97.9/88.8 & 96.8/82.0 & 98.8/90.1 & 99.2/92.2 \\
\textbf{Qwen2.5-14B-it} &96.4/74.9& 97.2/90.0 & 95.7/80.3 & 96.9/89.3 & 96.9/88.9 \\
\textbf{Qwen3-8B} &98.0/67.1 & 99.1/88.5 &98.7/73.1 &99.2/85.6 &98.4/89.7 \\
\textbf{internlm2-chat-7b} & 86.5/60.9 & 87.6/79.4 & 87.8/71.8 & 89.0/78.0 & 90.0/81.4 \\
\textbf{glm-4-9b-chat} & 96.4/67.7 & 97.8/84.6 & 95.8/79.5 & 99.3/87.6 & 98.7/89.6 \\
\textbf{Mistral-7B-it-v0.3} & 94.7/58.6 & 96.4/82.0 & 95.9/68.1 & 97.7/79.8 & 94.9/80.7\\
\textbf{Mistral-Nemo-it-2407} & 83.0/61.6 & 89.8/82.0 & 89.2/76.3 & 94.2/83.5 & 91.6/83.8 \\
\textbf{DeepSeek-V2-Lite-Chat} & 74.8/48.4 & 81.6/60.1 & 75.0/54.0& 78.8/55.6&77.7/58.6 \\
\textbf{Yi-1.5-9B-Chat-16K} & 76.8/56.7 & 85.1/80.2 & 78.9/58.0 & 82.6/69.8 & 87.8/76.2\\

\midrule
\textbf{Models} & \textbf{Industrials} & \textbf{Real Estate} &\textbf{Technology} & \textbf{Utilities} & \textbf{Average}\\
\midrule
\multicolumn{6}{c}{\textit{Private Models}} \\
\midrule
\textbf{GPT-5}  & 94.9/89.3 & 95.0/77.1 & 92.7/85.5 & 91.9/86.3 & 94.6/87.4   \\
\textbf{GPT-4}  & 95.4/88.7 & 96.3/87.2 & 93.8/90.4 & 91.5/91.7 & 95.9/89.5  \\
\textbf{Claude-3.5-Haiku}  & 96.2/93.4 & 93.2/87.1 & 95.5/93.6 & 94.4/92.4 & 97.1/91.0  \\
\textbf{Claude-4-Sonnet} & 93.9/83.4 & 87.4/70.5 & 91.1/81.5 & 92.9/76.5 & 90.6/79.5   \\

\midrule
\multicolumn{6}{c}{\textit{Open-Source Models}} \\
\midrule

\textbf{gemma-7b-it} & 97.7/89.6 & 98.2/88.4 & 97.4/90.1 & 99.1/88.0 & 97.6/87.6 \\
\textbf{gemma-2-9b-it} & 97.4/91.5 & 95.9/84.1 & 95.6/83.8 & 97.1/90.9 & 97.0/87.7 \\
\textbf{gemma-2-27b-it} & 97.4/89.9 & 97.1/80.5 & 97.1/90.0 & 98.6/87.1 & 97.5/86.9 \\
\textbf{Meta-Llama-3-8B-it} & 97.6/88.1  & 93.4/80.8  & 93.9/84.0  & 96.8/85.4  & 95.4/84.4  \\
\textbf{Meta-Llama-3.1-8B-it} & 94.1/88.5 & 91.9/84.8 & 91.2/85.8 & 94.0/84.7 & 92.4/84.5  \\
\textbf{Qwen2-7B-it} & 97.1/88.4 & 92.5/84.2 & 90.4/85.0 & 97.4/87.2 & 94.3/85.1 \\
\textbf{Qwen2.5-7B-it} & 98.1/90.3 & 98.5/90.2 & 98.8/93.1 & 99.2/91.0 & 98.5/89.1\\
\textbf{Qwen2.5-14B-it} &95.4/88.3&97.7/86.5&94.4/91.7&97.8/86.9&96.3/87.7 \\
\textbf{Qwen3-8B} &98.1/84.7 &98.0/82.3 &96.9/89.9&98.9/80.3 & 98.3/84.4 \\
\textbf{internlm2-chat-7b} & 90.4/81.7 & 85.9/78.7 & 88.9/74.0 & 92.2/80.7 & 89.1/77.5 \\
\textbf{glm-4-9b-chat} & 98.3/88.0 & 97.5/87.0 & 98.0/90.5 & 98.6/88.0 & 98.1/86.2 \\
\textbf{Mistral-7B-it-v0.3} & 97.4/82.6 & 94.2/74.5 & 96.4/83.3 & 97.6/80.5 & 96.4/78.8\\
\textbf{Mistral-Nemo-it-2407} & 94.1/84.4 & 92.4/78.9 & 87.2/80.2 & 92.7/81.6 & 90.8/80.7 \\
\textbf{DeepSeek-V2-Lite-Chat} &83.2/58.1&72.3/58.1& 71.8/59.1 & 82.5/59.3 & 78.0/57.3 \\
\textbf{Yi-1.5-9B-Chat-16K} & 86.7/71.3 & 78.2/65.5 & 78.7/68.1 & 82.2/65.8 & 83.4/69.7 \\

\bottomrule

\end{tabular}
\caption{Model Herding Score for contexts under two conditions: with and without Analyst Rating. For each cell, the left hand side of '/' is herding score under the case where analyst rating is included in report, the right hand side of '/' is herding score under the case where analyst rating is removed.}
\label{tab:herding_analyst}
\end{table*}

\begin{table*}[ht]
\centering
\resizebox{\textwidth}{!}{
\begin{tabular}{lcccccccccccc}
\hline
Model  &
\makecell{Comm.\\Services} &
\makecell{Cons.\\Defensive} &
Energy &
\makecell{Financial\\Services} &
Healthcare &
Industrials &
\makecell{Real \\Estate} &
Technology &
Utilities &
Avg \\
\hline
\multicolumn{11}{c}{\textit{Private Models}} \\
\hline
\textbf{GPT-5} & 50.00 & 39.61 & 32.20 & 26.70 & 28.49 & 29.80 & 38.64 & 32.19 & 42.65 & 33.54 \\
\textbf{GPT-4} & 64.24& 41.55 & 60.65 & 58.38& 51.24& 43.87&57.07&30.71&55.92& 48.77\\
\textbf{Claude-3.5-Haiku} &60.23 &39.34 &58.65 &54.97 & 48.12& 46.03& 54.62& 36.41&50.71 &45.98 \\
\textbf{Claude-4-Sonnet} & 52.06& 40.60& 37.09& 30.75& 33.40& 43.73& 35.45& 36.70& 40.02& 35.85\\
\hline
\multicolumn{11}{c}{\textit{Open-Source Models}} \\
\hline 
\textbf{Qwen2.5-7B-it} & 41.33  & 22.96 & 41.19 & 33.96 & 33.01 & 34.07 & 32.96 & 14.77 & 34.08 & 30.33 \\
\textbf{Qwen2-7B-it} & 52.65 & 28.47 & 42.29 & 40.66 & 31.10 & 22.02 & 36.48 & 20.57 & 33.28 & 31.69 \\
\textbf{Qwen2.5-14B-it} & 65.20 &  38.89 & 60.79 & 62.80 & 50.56 &33.66 & 54.55 & 35.17 & 40.00 & 46.64 \\
\textbf{Qwen3-8B} & 46.18 & 31.53 & 64.72 & 45.79 & 44.95 & 26.17 & 45.57 & 17.43 & 38.25 & 36.89 \\
\textbf{gemma-2-27b-it} & 68.76 &  43.25 & 67.64 & 62.00 & 53.81 & 42.20 & 58.22 & 23.07 & 55.51 & 48.78 \\
\textbf{gemma-2-9b-it} & 48.01 &  19.68 & 49.25 & 29.15 & 30.06 & 16.61 & 33.18 & 17.74 & 31.39 & 27.14 \\
\textbf{gemma-7b-it} & 19.93 &  11.62 & 20.98 & 7.91 & 9.95 & 12.59 & 9.34 & 7.57 & 12.98 & 11.48 \\
\textbf{internlm2-chat-7b} & 39.91 & 30.46 & 39.55 & 35.28 & 29.70 & 25.08 & 29.55 & 25.23 & 26.25 & 30.34 \\
\textbf{Mistral-7B-it-v0.3} & 69.90 &  59.04 & 72.96 & 69.77 & 68.35 & 55.24 & 65.15 & 37.74 & 66.38 & 60.42 \\
\textbf{Mistral-Nemo-it-2407} & 50.17 &  30.00 & 40.52 & 37.81 & 29.03 & 26.50 & 30.96 & 22.45 & 29.18 & 31.28 \\
\textbf{Yi-1.5-9B-Chat-16K} & 32.43 &  27.24 & 38.50 & 34.63 & 33.80 & 32.40 & 34.01 & 30.85 & 34.91 & 32.80 \\
\textbf{DeepSeek-V2-Lite-Chat} & 57.09 &  56.80 & 55.54 & 60.69 & 56.60 & 55.52 & 58.87 & 47.52 & 59.86 & 55.87 \\
\textbf{Meta-Llama-3-8B-it} & 64.40 &  42.78 & 50.41 & 47.95 & 48.37 & 39.40 & 46.95 & 56.06 & 47.28 & 48.57 \\
\textbf{Meta-Llama-3.1-8B-it} & 34.80 &  17.60 & 26.63 & 19.35 & 19.17 & 12.08 & 22.78 & 14.12 & 19.66 & 18.78 \\
\textbf{glm-4-9b-chat} & 70.99 &  46.61 & 62.56 & 53.73 & 53.65 & 45.29 & 51.91 & 39.88 & 58.56 & 51.30 \\
\hline
\end{tabular}}
\caption{Model Herding Score given contexts where original analyst rating is randomly replaced with a 'fake' rating which is different from the original one.}
\label{tab:herding_fake}
\end{table*}

\subsection{Results}

To better demonstrate our findings, we organize our work around three key research questions:

\textbf{RQ1:} How LLM will align with analyst in providing investment rating when analyst rating is removed in context?

\textbf{RQ2:} How LLM will align with 'fake' rating in providing investment rating even when the 'fake' rating is contradictory with the rest of context?

\textbf{RQ3:} How "LLM herding" can affect LLM performance in stock trading? Can LLMs exceed sophisticated analysts when independently making investment decision?

Table \ref{tab:herding_analyst} reveals the herding behavior for each model under two different conditions: with or without analyst rating in context. We find that the inclusion of analyst ratings significantly strengthens model herding behavior across nearly all categories and models. Specifically, the results suggest that when analyst ratings are not explicitly stated, private models (like GPT-5, GPT-4, and Claude variants) generally align with analyst ratings more than open-source models, with average herding scores in the mid-80s to low-90s. For instance, GPT-5 achieves an average of 87.4\%, while GPT-4 follows closely at 89.5\%. Open-source models, by contrast, display more variability, with averages ranging from the mid-60s (e.g., Yi-1.5-9b-Chat at 69.7\%) to the mid-80s (e.g., Llama-3-8b-it at 84.4\%). This is intuitive because private models such as GPT, Claude tend to have way more parameters to capture contextual and semantic information, making them align with analysts better even without explicit rating guidance. In contrast, table \ref{tab:herding_analyst} also shows a striking shift once analyst ratings are provided. Across both private and open-source models, herding scores rise sharply, with many models reaching averages above 90\%. GPT-5 and GPT-4, for instance, jump to 94.6\% and 95.9\%, respectively, while even mid-performing open-source models like gemma-7b-it and Mistral-7B-it-v0.3 achieve scores exceeding 96\%. Notably, models that previously lagged in case of no analyst rating (such as Yi-1.5-9b-Chat or Qwen2-7b-it) demonstrate significant improvements, often surpassing 90\% in most categories. This suggests that human ratings act as a strong guiding signal, discouraging models' independent thinking and reducing divergence in model outputs and encouraging greater alignment across sectors. A case study is provided in the appendix. One noticeable thing is that the increased model herding behavior from adding explicit analyst ratings are not uniform across sectors. For example, sectors like Financial Services and Healthcare already see relatively high alignment (often above 85–90\%), while sectors like Communication Services and Utilities show greater variability. When analyst ratings are provided, these weaker sectors benefit disproportionately, with scores consolidating above 90\% for nearly all models. This highlights that external expert input (analyst ratings) particularly enhances agreement in areas where model consensus is weaker. 
\\
\begin{table*}[ht]
\centering
\scriptsize 
\setlength{\tabcolsep}{3pt}
\renewcommand{\arraystretch}{0.9}
\begin{tabular}{lccccc}
\toprule
\textbf{Models} & 
\textbf{Communication Services} & 
\textbf{Consumer Defensive} & \textbf{Energy} & \textbf{Financial Services} & \textbf{Healthcare} \\

\midrule
\multicolumn{6}{c}{\textit{Private Models}} \\
\midrule
\textbf{GPT-5}  & 33.0/33.2  & 34.8/35.7  & 35.1/33.3  & 31.9/32.6 &30.1/31.1  \\
\textbf{GPT-4}  & 31.6/32.8  & 33.8/35.0  &  34.1/32.1 & 32.4/33.5& 31.3/33.0  \\
\textbf{Claude-3.5-Haiku}  &31.1/32.4  &34.0/35.0 & 34.3/33.4 & 31.5/34.2& 30.1/30.8  \\
\textbf{Claude-4-Sonnet} &32.5/33.1   & 33.9/34.8  & 35.2/34.6  & 33.1/34.5&31.1/31.9   \\

\midrule
\multicolumn{6}{c}{\textit{Open-Source Models}} \\
\midrule

\textbf{gemma-7b-it} &30.3/24.3  &32.5/27.2 & 32.0/26.0 &34.1/29.7  &31.2/25.9 \\
\textbf{gemma-2-9b-it} &29.9/29.1  &33.5/30.1  &33.4/25.9  &35.1/29.0 & 31.5/27.7 \\
\textbf{gemma-2-27b-it} &27.9/28.8 &33.9/33.8  &32.6/32.6 &34.1/33.1 &31.2/30.3 \\
\textbf{Meta-Llama-3-8B-it} &32.2/26.9  & 34.2/28.8 &31.3/28.1 &34.1/28.4 & 31.1/27.1 \\
\textbf{Meta-Llama-3.1-8B-it} &33.6/25.4  &33.3/29.1   &34.8/29.7   &34.3/29.0 &31.4/25.9  \\
\textbf{Qwen2-7B-it} &30.4/26.3 &34.6/28.1 &32.2/26.4 &33.8/27.8 &30.7/25.8 \\
\textbf{Qwen2.5-7B-it} &29.0/33.6 &33.4/32.5 &32.2/31.1 &33.2/33.5 &31.2/31.4 \\
\textbf{Qwen2.5-14B-it} &31.0/34.8 &31.7/35.3 &34.0/33.7 &32.2/33.2 &31.4/30.5 \\
\textbf{Qwen3-8B} &29.4/33.9 &33.9/34.5 &31.9/31.1 &33.8/35.1 &31.7/32.6 \\
\textbf{internlm2-chat-7b} &30.7/24.9 &36.0/26.6 &33.5/28.2 &34.5/29.3 &30.5/25.5 \\
\textbf{glm-4-9b-chat} &30.1/24.9 &32.8/26.3 &33.0/26.9 &34.3/28.7 &31.3/24.7 \\
\textbf{Mistral-7B-it-v0.3} & 30.4/35.0&33.4/33.5 &33.0/32.9 &34.1/34.1 &30.7/32.7\\
\textbf{Mistral-Nemo-it-2407} &31.7/33.6 &34.8/35.2 &33.9/31.1 &34.4/34.5 &31.0/33.4 \\
\textbf{DeepSeek-V2-Lite-Chat} &32.0/34.8 &34.1/35.6 &32.1/32.4 &35.1/35.5 &33.9/32.0 \\
\textbf{Yi-1.5-9B-Chat-16K} &32.9/28.5 &36.2/29.5 &34.7/28.5 &36.5/30.4 & 32.9/28.1\\
Analyst & 29.0 & 33.8 & 32.4 & 33.8 & 31.6 \\ 

\midrule
\textbf{Models} & \textbf{Industrials} & \textbf{Real Estate} &\textbf{Technology} & \textbf{Utilities} & \textbf{Average}\\
\midrule
\multicolumn{6}{c}{\textit{Private Models}} \\
\midrule
\textbf{GPT-5}  & 31.2/34.0  & 28.7/28.8  & 33.2/34.7 &35.1/37.4 & 33.0/33.6   \\
\textbf{GPT-4}  & 30.9/33.1  & 28.9/29.9  & 33.0/33.6  & 33.1/35.9& 33.0/33.9  \\
\textbf{Claude-3.5-Haiku}  & 30.9/32.5 & 27.6/27.9 & 32.8/33.1 & 32.5/34.5& 32.6/33.5  \\
\textbf{Claude-4-Sonnet} & 30.9/33.9  & 29.1/29.6  & 32.5/34.5  & 35.0/35.8 & 33.0/33.8   \\

\midrule
\multicolumn{6}{c}{\textit{Open-Source Models}} \\
\midrule

\textbf{gemma-7b-it} & 31.8 / 27.9 &
30.5 / 24.9 &
36.6 / 26.3 &
33.0 / 24.2 & 32.9 / 26.7 \\
\textbf{gemma-2-9b-it} & 31.7 / 28.4 &
32.7 / 28.0 &
36.8 / 26.1 &
32.1 / 25.9& 33.0 / 27.8  \\
\textbf{gemma-2-27b-it} & 31.4 / 31.2 &
32.1 / 30.7 &
36.1 / 37.2 &
32.1 / 32.6 & 32.6 / 32.3 \\
\textbf{Meta-Llama-3-8B-it} & 32.2 / 28.4 &
32.1 / 26.9 &
37.0 / 26.4 &
32.5 / 26.0& 33.4 / 27.5  \\
\textbf{Meta-Llama-3.1-8B-it} & 31.8 / 28.1 &
31.8 / 25.1 &
36.6 / 26.3 &
32.3 / 24.5 & 33.5 / 27.2  \\
\textbf{Qwen2-7B-it} & 32.1 / 28.9 &
30.3 / 25.7 &
36.2 / 26.8 &
34.0 / 25.9& 33.1 /27.1 \\
\textbf{Qwen2.5-7B-it} & 32.1 /32.8 &
31.3 /33.7  &
35.4 /36.6  &
33.3 /31.2  & 32.7/33.2\\
\textbf{Qwen2.5-14B-it} & 33.2 / 32.4 &
31.5 / 30.8 &
33.2 / 35.1 &
30.9 / 31.7& 32.3 / 33.1 \\
\textbf{Qwen3-8B} & 32.2 / 33.5 &
31.1 / 30.8 &
36.8 / 37.4 &
33.0 / 33.5 & 33.1 / 34.2 \\
\textbf{internlm2-chat-7b} &32.2 / 29.0 &
31.5 / 26.4 &
36.1 / 26.5 &
32.8 / 24.9 &33.4 / 27.1 \\
\textbf{glm-4-9b-chat} & 32.4 / 27.9 &
30.2 / 25.3 &
36.9 / 26.6 &
31.6 / 23.4 & 33.1 / 26.4 \\
\textbf{Mistral-7B-it-v0.3} & 32.0 / 33.5 &
29.5 / 32.8 &
37.2 / 37.5 &
32.7 / 32.6 & 33.1 / 34.1\\
\textbf{Mistral-Nemo-it-2407} & 32.4 / 34.0 &
30.9 / 32.9 &
37.8 / 37.5 &
32.6 / 31.9 & 33.7 / 34.3 \\
\textbf{DeepSeek-V2-Lite-Chat} &32.6/ 33.9 &
33.4 / 38.7 &
35.1 / 34.9 &
35.1 / 36.0 & 33.9/ 34.6 \\
\textbf{Yi-1.5-9B-Chat-16K} & 33.3 / 30.0 &
35.1 / 26.4 &
33.1 / 24.1 &
35.0 / 28.9 &34.5 / 28.7 \\
\textbf{Analyst} & 32.0 & 31.1 & 36.5 & 33.0 & 33.1 \\
\bottomrule

\end{tabular}
\caption{Model Accuracy for contexts under two conditions: with and without Analyst Rating. For each cell, the left hand side of '/' is herding score under the case where analyst rating is included in report, the right hand side of '/' is herding score under the case where analyst rating is removed.}
\label{tab:accuracy_wo_analyst}
\end{table*}

\begin{table*}[ht]
\centering
\resizebox{\textwidth}{!}{
\begin{tabular}{lcccccccccc}
\hline
Model  &
\makecell{Comm.\\Services} &
\makecell{Cons.\\Defensive} &
Energy &
\makecell{Financial\\Services} &
Healthcare &
Industrials &
\makecell{Real \\Estate} &
Technology &
Utilities &
Avg \\
\hline
\multicolumn{11}{c}{\textit{Private Models}} \\
\hline
\textbf{GPT-5} & 33.56 & 37.78 & 34.86 & 32.92 & 31.52 & 33.91 & 33.59 &  34.74& 35.66 & 34.16 \\
\textbf{GPT-4} & 36.27 & 36.56 & 33.82 & 34.38 & 35.87 &35.65 & 36.11 &36.01 & 39.37& 35.88 \\
\textbf{Claude-3.5-Haiku} &35.31 & 37.12& 35.04& 33.43& 30.98& 33.74& 35.25& 35.03&34.61 & 34.99\\
\textbf{Claude-4-Sonnet} &35.79 & 37.54& 33.96& 35.13& 34.69& 35.63&36.86 &35.47 & 38.98& 35.70\\
\hline
\multicolumn{11}{c}{\textit{Open-Source Models}} \\
\hline
\textbf{Qwen2-7B-it} & 32.63 & 33.12 & 34.79 & 33.03 & 31.84 & 33.12 & 38.30 & 36.57 & 33.14 & 33.82 \\
\textbf{gemma-2-9b-it} & 38.11 & 35.77 & 34.27 & 34.52 & 33.13 & 32.68 & 35.78 & 36.03 & 35.35 & 34.71 \\
\textbf{gemma-7b-it} & 33.56 & 31.87 & 32.20 & 35.04 & 31.08 & 32.17 & 35.86 & 35.77 & 29.81 & 33.12 \\
\textbf{glm-4-9b-chat} & 33.28 & 31.27 & 34.81 & 32.45 & 30.42 & 31.50 & 35.03 & 35.13 & 30.55 & 32.53 \\
\textbf{Mistral-Nemo-It-2407} & 36.50 & 33.50 & 33.17 & 32.74 & 33.83 & 34.64 & 36.80 & 35.89 & 36.29 & 34.58 \\
\textbf{Mistral-7B-It-v0.3} & 38.81 & 34.05 & 36.64 & 35.26 & 37.89 & 34.84 & 38.38 & 35.33 & 36.23 & 35.99 \\
\textbf{DeepSeek-V2-Lite-Chat} & 37.09 & 32.17 & 35.07 & 33.26 & 36.29 & 38.06 & 35.04 & 35.20 & 35.24 & 35.30 \\
\textbf{Meta-Llama-3-8B-it} & 36.56 & 35.20 & 35.02 & 35.19 & 33.33 & 33.87 & 38.07 & 34.81 & 35.71 & 34.91 \\
\textbf{Meta-Llama-3.1-8B-it} & 33.05 & 32.24 & 33.82 & 33.04 & 31.05 & 31.48 & 32.06 & 33.71 & 33.43 & 32.58 \\
\textbf{internlm2-chat-7b} & 36.01 & 32.76 & 36.54 & 32.54 & 32.70 & 33.10 & 34.49 & 35.25 & 32.09 & 33.69 \\
\textbf{gemma-2-27b-it} & 36.94 & 35.97 & 35.66 & 35.73 & 35.36 & 35.09 & 42.09 & 36.96 & 36.52 & 36.23 \\
\textbf{Qwen2.5-14B-it} & 32.59 & 33.89 & 34.28 & 35.92 & 35.37 & 32.17 & 38.46 & 35.63 & 33.67 & 34.60 \\
\textbf{Qwen2.5-7B-it} & 33.59 & 31.77 & 34.19 & 33.66 & 33.39 & 33.49 & 36.13 & 34.93 & 32.32 & 33.64 \\
\textbf{Yi-1.5-9B-Chat-16K} & 36.44 & 36.30 & 35.77 & 36.28 & 39.98 & 37.61 & 40.66 & 35.73 & 43.22 & 37.67 \\
\textbf{Qwen3-8B} & 36.39 & 35.27 & 36.72 & 35.48 & 37.02 & 35.59 & 39.90 & 35.91 & 40.20 & 36.49 \\
\textbf{Analyst} & 28.98 & 33.81 & 32.36 & 33.80 & 31.60 & 31.96 & 31.06 & 36.49 & 32.95 & 33.08 \\
\hline
\end{tabular}}
\caption{Model investment rating prediction accuracy given contexts with all human opinions removed. The ground truth label has three categories 'bullish', 'bearish' and 'neutral'. The calculation details of ground truth label can be found in appendix.}
\label{tab:accuracy_wo_opinion}
\end{table*}

Table \ref{tab:herding_fake} displays experiment results for model herding score when perturbing original analyst rating statement shown at the beginning of analyst report. The true analyst rating is deliberately replaced by another different fake rating which may be contradictory to the main analysis content. The results suggest that there is a large variation in herding fake rating between models, ranging from 10\% to 60\%. The average herding score across all models is approximately 30\%, indicating that LLMs, even state-of-the-art systems such as GPT-5 and GPT-4 remain susceptible to inheriting unsupported human biases embedded in the input. This suggests that substantial potential for model herding persists, even among the most advanced architectures. Additionally, the degree of herding is not strongly correlated with model size. 
\\

Table \ref{tab:accuracy_wo_analyst} reports LLM performance in providing stock investment ratings based solely on analyst reports using real time stock return as ground-truth label, with or without access to the analysts’ explicit investment ratings. When analyst ratings are provided, nearly all models achieve accuracy levels close to those of human analysts, indicating strong herding behavior. The average analyst accuracy is approximately 33\% and the accuracy of most models falls between 32\% - 34\%. In contrast, when the first sentences that contain explicit analyst ratings are removed, model accuracy diverges substantially, ranging from -6\% to +2\% relative to analysts. The removal of first rating sentence tend to have least effects on the performance of private models like GPT-5, GPT-4, Claude-3.5-Haiku and Claude-4-Sonnet, the change of accuracy for each model is less than 1\%. In contrast, several open-source models fall notably below analyst performance, including gemma-2-9b-it (-6\%), gemma-7b-it (-6\%), Meta-Llama-3-8B-Instruct (-6\%), Meta-Llama-3.1-8B-Instruct (-6\%), internlm2-chat-7b (-6\%), Yi-1.5-9B-Chat (-6\%), and glm-4-9b-chat (-6\%).
At the same time, several models outperform analysts without access to analyst ratings, such as Mistral-Nemo-it-2407 (+1\%), DeepSeek-V2-Lite-Chat (+1\%), Mistral-7B-it-v0.3 (+1\%), and Qwen3-8B (+1\%). The large drop in accuracy when analyst ratings are removed further illustrates herding behavior: models rely heavily on human opinions to perform well. Although alignment with analyst views improves apparent accuracy for some models, it also shows that models lack independent reasoning, as their performance does not surpass human analysts. Across industries, LLMs generally achieve higher accuracy in sectors such as Financial Services, Utilities, and Technology.

\section{Alleviating model herding via bias awareness}

\subsection{Lexicon Filtering}

We propose a method to mitigate model herding in discrete decision-making by filtering human opinions from the input context, thereby encouraging models to reason independently. Although we remove the explicit analyst rating from the first sentence of each report, the remaining text often still contains numerous sentences expressing explicit or implicit subjective judgments, which can induce implicit herding behavior in LLMs. To identify such potentially biased statements, we employ the Multi-Perspective Question Answering (MPQA) Subjectivity Lexicon, a widely used linguistic resource for detecting opinion-bearing and subjective expressions in text ~\citep{wiebe2005annotating}. The lexicon comprises several thousand lexical items annotated with rich subjectivity metadata—including polarity (positive, negative, neutral, or both), subjectivity strength (“strongsubj” or “weaksubj”), and part-of-speech categories (noun, verb, adjective, adverb). In our implementation, we perturb each analyst report by removing sentences that contain any lexical item labeled as strongsubj in MPQA. An example is provided in the appendix. As reported in Table~\ref{tab:accuracy_wo_opinion}, the method proves effective for both private models and open-source models. The private models like GPT-4, Claude-4-Sonnet improves 2\%, while GPT-5 only improves 0.5\%. In contrast, for open-source models, the accuracy of LLM-generated investment ratings mostly improves significantly after excluding all opinionated content, with gains of approximately 2–4 percentage points relative to the case where only the first rating sentence is excluded. Most importantly, some models including Qwen3-8B, DeepSeek-V2-Lite-Chat and Meta-Llama-3-8B-It perform even better than analysts as well as cutting-edge complex models like GPT-5. These results suggest that filtering biased subjective expressions enhances LLM performance by reducing reliance on human opinions embedded in the text and light open-source models have great potentials if properly prompted which can help save costs. 

\subsection{Reducing model herding via Direct Preference Optimization (DPO)}

To address the limitations of static prompting and data filtering that do not reflect real-world settings where models are exposed to raw, biased context, we implement a Direct Preference Optimization (DPO) framework. This workflow transforms the debiasing task from a sequence-level instruction into an architectural preference. We first construct a triplet dataset consisting of a financial context (input $x$), an 'Independent' response ($y_w$), and a 'Herded' response ($y_l$). The input $x$ is original analyst report that contain biased ratings or opinions. The herded response ($y_l$) is generated by prompting an LLM to reason following the analyst logic through keeping analyst opinions in the context, effectively simulating the herding bias. Conversely, the independent response ($y_w$) is generated by providing the model with the context without analyst opinions, adding actual market ground-truth rating and instructing it to derive a reasoning chain that leads to that outcome while explicitly disregarding the analyst opinions. By optimizing the DPO objective that maximizes the log-likelihood ratio between the preferred and rejected responses, the model internalizes a skepticism toward high-consensus human signals and focus more on firm fundamental evidence. Since we cannot fine-tune close models, we select a representative set of open models including Qwen3-8B, Qwen2.5-7B-it and llama-3 which are mostly used in prior literature for fine-tuning. The results in table \ref{tab:accuracy_dpo} suggest our fine-tuning method is effective in mitigating model herding and improving prediction accuracy.

\section{Conclusion}

In this study, we present a comprehensive investigation into the decision-making behavior of large language models (LLMs) in the context of financial investment analysis. By introducing the Fin-Bias benchmark, we evaluate whether LLMs exhibit herding behavior when exposed to potentially biased human opinions or even fake opinions embedded in long-form analyst reports. Our experiment results demonstrate that LLMs are indeed susceptible to herding, with significantly higher alignment to analyst ratings when such ratings are present in the input.  These findings challenge the assumption that larger or more advanced models inherently reason more independently, and
468 highlight the importance of carefully curating inputs to encourage unbiased and grounded financial reasoning. We further demonstrate the necessity
to remove implicit human bias to strengthen LLM
independent reasoning. Overall, our work underscores the need for future LLM development to prioritize independent judgment over mere reflection of dominant human narratives, especially in opinion-driven domains.  

\section{Limitations}

First, Fin-Bias currently focuses on single-agent financial decision-making task without addressing LLM herding in multi-agent framework. Second, the data we use come from financial analysts which represent most investors but not the whole. Third, we only study LLM herding behavior in finance domain which may or may not be extended to other domains.


\bibliography{custom}

\appendix

\section{Appendices}

\begin{table*}[ht]
\centering
\resizebox{\textwidth}{!}{
\begin{tabular}{lcccccccccc}
\hline
Model  &
\makecell{Comm.\\Services} &
\makecell{Cons.\\Defensive} &
Energy &
\makecell{Financial\\Services} &
Healthcare &
Industrials &
\makecell{Real \\Estate} &
Technology &
Utilities &
Avg \\
\hline
\textbf{Qwen2-7B-it} & 34.72 & 35.45 & 35.83 & 34.15 & 32.90 & 34.58 & 39.08 & 37.10 & 36.33 & 35.15 \\
\textbf{Meta-Llama-3-8B-it} & 37.43 & 37.88 & 38.52& 38.67 & 35.21 & 34.95 & 40.15 & 37.02 & 36.29 & 37.09 \\
\textbf{Qwen3-8B} & 37.58 & 37.34 & 38.70 & 36.73 & 38.90 & 36.02 & 40.15 & 38.69 & 41.33 & 38.23 \\

\hline
\end{tabular}}
\caption{Model investment rating prediction accuracy given contexts using DPO post-training. The ground truth label has three categories 'bullish', 'bearish' and 'neutral'. The calculation details of ground truth label can be found in appendix.}
\label{tab:accuracy_dpo}
\end{table*}

\subsection{Ground-Truth Label Calculation Details}

Specifically, we employ a quantile-based portfolio classification derived from the three-month (approximately 60 trading days) cumulative abnormal return (CAR) following the report issuance date. This design reflects the fact that analyst reports primarily target a firm’s medium- to long-term performance rather than short-horizon fluctuations in daily returns. On average, analysts update their forecasts when firm issuing quarterly report, so we believe 60 trading days is average forecast horizons. In particular, for each firm, given a report issued on YYYY/MM/DD, we compute the 60-day abnormal return ($\alpha$) using the market model. 

The market model is a standard asset pricing framework used to decompose an asset’s return into a market-driven component and an asset-specific abnormal component (alpha).

\paragraph{Model specification.}
For asset $i$ at time $t$, the market model is defined as:
\begin{equation}
R_{i,t} = \alpha_i + \beta_i R_{m,t} + \varepsilon_{i,t},
\end{equation}
where $R_{i,t}$ denotes the return of firm $i$, $R_{m,t}$ denotes the return of the market portfolio, $\alpha_i$ represents the abnormal return (alpha), $\beta_i$ measures the sensitivity of firm $i$ to market movements, which can be seen as risk and $\varepsilon_{i,t}$ is an idiosyncratic error term satisfying
\begin{equation}
\mathbb{E}[\varepsilon_{i,t}] = 0, \quad \mathrm{Var}(\varepsilon_{i,t}) = \sigma^2_{\varepsilon_i}.
\end{equation}

\paragraph{Estimation of alpha.}
The parameters $\alpha_i$ and $\beta_i$ are estimated using ordinary least squares (OLS) over an estimation window $t = 1, \ldots, T$. The OLS estimator of $\beta_i$ is given by:
\begin{equation}
\hat{\beta}_i =
\frac{\sum_{t=1}^{T} (R_{m,t} - \bar{R}_m)(R_{i,t} - \bar{R}_i)}
{\sum_{t=1}^{T} (R_{m,t} - \bar{R}_m)^2},
\end{equation}
where
\begin{equation}
\bar{R}_i = \frac{1}{T}\sum_{t=1}^{T} R_{i,t}, \quad
\bar{R}_m = \frac{1}{T}\sum_{t=1}^{T} R_{m,t}.
\end{equation}

The estimated alpha is then computed as:
\begin{equation}
\hat{\alpha}_i = \bar{R}_i - \hat{\beta}_i \bar{R}_m.
\end{equation}

A positive value of $\alpha_i$ indicates that the asset generates abnormal returns beyond what is explained by market movements, whereas a negative value suggests underperformance relative to the market benchmark. Because investors such as hedge funds are fundamentally concerned with risk-adjusted abnormal returns rather than raw buy-and-hold performance, cumulative abnormal return provides a more economically meaningful basis for defining true investment ratings. \textbf{Cumulative Abnormal Return} measures the total alpha change of an investment over time by summing daily alpha, shown in Equation 7. Higher values indicate better strategy effectiveness.

\begin{equation}
CAR = \sum\hat{\alpha}_i = \sum(\bar{R}_i - \hat{\beta}_i \bar{R}_m)
\end{equation}\label{car}

After computing each report’s 60-day cumulative abnormal return, we classify investment outcomes using a year-specific, quantile-based approach inspired by long-short portfolio method widely documented in finance literature. For all reports issued in year t, we sort the post-report cumulative abnormal returns and assign ratings based on their relative positions within the annual distribution. Specifically, observations in the upper 30\% quantile are labeled 'bullish', those in the lower 30\% quantile are labeled 'bearish', and the remaining middle 40\% are categorized as 'neutral'. This procedure ensures that classifications are comparable across years and are not distorted by time-varying market conditions.

\subsection{Sample Original Analyst Report}

The analyst reports contain analysts' deep research for a specific firm's future stock price. They collect information from different sources ranging from financial statements, MD\&A to earnings conference calls and summarize multiple factors/events influencing company stock return, including macroeconomic situation, industry situation as well as firm fundamentals\& strategy in a professional way, which is of higher quality than stock news \& tweets. The reports tend to provide an investment rating(bullish/neutral/bearish) in the first sentence. Here is an example analyst report for oracle, which is shown in Figure \ref{fig:oracle-report}. 

\subsection{Sample data perturbation}

Figure \ref{fig:oracle-report-no-rating} and Figure \ref{fig:oracle-report-fake} give examples on perturbing original analyst report using two different methods: remove the first sentence which indicates the explicit analyst rating and replace the original investment rating with a 'fake' rating. The only difference between figure \ref{fig:oracle-report}, \ref{fig:oracle-report-no-rating} and \ref{fig:oracle-report-fake} lies in the first sentence. Figure \ref{fig:oracle-report-object} displays the same sample analyst report but with sentences containing human opinions removed using the method documented in section 5. 

\begin{figure*}
\centering
\begin{tcolorbox}[colback=teal!5, colframe=teal, title= Original Analyst Report Sample]
    
\textbf{We are maintaining our HOLD rating on Oracle Corp. (NGS: ORCL).} We have a sense of relief that the Oracle/TikTok deal is much more in line with a commercial transaction than a merger. Rumors swirling around the deal at one point had Oracle acquiring TikTok. In the actual deal, Oracle locks in TikTok's growing business for its small Oracle Cloud service as it gets a stake in a high-growth asset which could one day file for an IPO. Our biggest fears were around Oracle acquiring ownership in a business that had no strategic fit in an area, social media, in which it has zero experience. While it is doubtful that the deal will actually accomplish true security when it comes to TikTok's influence on U.S. social media, we think Oracle should be capable of fulfilling its commitments on U.S. user data security.
\\
....
\\
Oracle Corp. is the world's largest independent enterprise software company, with annual revenues of \$39 billion. Its software products include database, middleware, application and cloud-based software designed for general business purposes and for specific industries. In addition, Oracle provides product upgrades, maintenance releases and patches through license update agreements. It also offers product support through the cloud, internet and global support centers. Oracle also provides server hardware through its acquisition of Sun Microsystems in 2010.
\end{tcolorbox}
\caption{A sample analyst report used as input to the model.}
\label{fig:oracle-report}
\end{figure*}

\begin{figure*}
\centering
\begin{tcolorbox}[colback=teal!5, colframe=teal, title= Analyst Report without rating Sample]
    
We have a sense of relief that the Oracle/TikTok deal is much more in line with a commercial transaction than a merger. Rumors swirling around the deal at one point had Oracle acquiring TikTok. In the actual deal, Oracle locks in TikTok's growing business for its small Oracle Cloud service as it gets a stake in a high-growth asset which could one day file for an IPO. Our biggest fears were around Oracle acquiring ownership in a business that had no strategic fit in an area, social media, in which it has zero experience. While it is doubtful that the deal will actually accomplish true security when it comes to TikTok's influence on U.S. social media, we think Oracle should be capable of fulfilling its commitments on U.S. user data security.
\\
....
\\
Oracle Corp. is the world's largest independent enterprise software company, with annual revenues of \$39 billion. Its software products include database, middleware, application and cloud-based software designed for general business purposes and for specific industries. In addition, Oracle provides product upgrades, maintenance releases and patches through license update agreements. It also offers product support through the cloud, internet and global support centers. Oracle also provides server hardware through its acquisition of Sun Microsystems in 2010.
\end{tcolorbox}
\caption{A sample analyst report without rating as input to the model.}
\label{fig:oracle-report-no-rating}
\end{figure*}

\begin{figure*}
\centering
\begin{tcolorbox}[colback=teal!5, colframe=teal, title= Analyst Report with fake rating Sample]
    
\textbf{We are maintaining our SELL rating on Oracle Corp. (NGS: ORCL).} We have a sense of relief that the Oracle/TikTok deal is much more in line with a commercial transaction than a merger. Rumors swirling around the deal at one point had Oracle acquiring TikTok. In the actual deal, Oracle locks in TikTok's growing business for its small Oracle Cloud service as it gets a stake in a high-growth asset which could one day file for an IPO. Our biggest fears were around Oracle acquiring ownership in a business that had no strategic fit in an area, social media, in which it has zero experience. While it is doubtful that the deal will actually accomplish true security when it comes to TikTok's influence on U.S. social media, we think Oracle should be capable of fulfilling its commitments on U.S. user data security.
\\
....
\\
Oracle Corp. is the world's largest independent enterprise software company, with annual revenues of \$39 billion. Its software products include database, middleware, application and cloud-based software designed for general business purposes and for specific industries. In addition, Oracle provides product upgrades, maintenance releases and patches through license update agreements. It also offers product support through the cloud, internet and global support centers. Oracle also provides server hardware through its acquisition of Sun Microsystems in 2010.
\end{tcolorbox}
\caption{A sample analyst report with 'fake' rating as input to the model.}
\label{fig:oracle-report-fake}
\end{figure*}

\begin{figure*}
\centering
\begin{tcolorbox}[colback=teal!5, colframe=teal, title= Analyst Report without human opinions Sample]

Founder and CTO Larry Ellison has identified two critical Oracle products that will determine the company's future: its cloud ERP application software and its autonomous cloud database (note the word 'cloud' in each of these products). In  addition,  smaller  niche  competitors  such  as Salesforce.com  and  Workday  have  targeted  specific software-as-a-service applications. ORCL trades near its historical discount to peers. TikTok has already been called out for tailoring its algorithm to Chinese government policies, as any Chinese company is required to do. Oracle had earlier confirmed that it was 'part of a proposal submitted by ByteDance to the Treasury Department' on September 14. ByteDance is the owner of virally popular social media site TikTok, which was under a September 20 deadline for a sale to a U.S. company; that deadline has now been extended, at least, until September 27 to accommodate the Oracle deal. However, China, where ByteDance is domiciled, passed technology export restrictions which in effect prevented a sale. Oracle Corp. is the world's largest independent enterprise software company, with annual revenues of \$39 billion. In addition, Oracle provides product upgrades, maintenance releases and patches through license update agreements. Oracle also provides server hardware through its acquisition of Sun Microsystems in 2010.The company has over 100,000 employees. About 48\% of revenues come from overseas. ORCL has traded between $39 and $63 over the past year and is currently at the high end of that range. ORCL shares are up 15\% on a total-return basis year-to-date, compared to a 2\% gain for the S\&P and a 22\% gain for the S\&P Information Technology Index. Oracle's forward enterprise value/EBITDA multiple of 10.5 is 27\% below the peer average, near the average discount of 26\% over the past two years."

\end{tcolorbox}
\caption{A sample analyst report without human opinions as input to the model.}
\label{fig:oracle-report-object}
\end{figure*}

\subsection{Case Study on LLM herding towards human opinions}

Here is sample case study for LLM outputs (Meta-llama-3-It) given contexts in two different conditions: with analyst rating and without analyst rating. Figure \ref{fig:BAC-report} shows a context With analyst rating and Figure \ref{fig:BAC-LLM-output-with-rating} demonstrates the LLM output given that context using Chain-of-thought prompt. We can see LLM tend to herd analyst investment rating reflected in reasoning. In Figure \ref{fig:BAC-report-wo-rating}, the analyst rating is removed in context. Figure \ref{fig:BAC-LLM-output-wo-rating} shows LLM outputs a different rating and provides relatively independent thinking.

\begin{figure*}
\centering
\begin{tcolorbox}[colback=teal!5, colframe=teal, title= Context]
    
\textbf{We are maintaining our BUY rating on Bank of America Corp. (NYSE: BAC) and target price of \$30 following 1Q results.} The quarter included interest margin contraction as lower rates take hold, and a sharply higher credit loss provision as the bank prepares for delinquencies from the fallout of measures taken to contain the coronavirus. We look for continued pressure on revenues (from reduced loan volumes, year-over-year margin contraction and lower fee-based income) in the second half of 2020 to keep earnings below the recent run rate. Looking beyond near-term challenges, management continues to focus on what it terms 'responsible growth.' We believe this may be seen in the company's ability to expand its loan portfolio without taking on too much credit risk, and to maintain balanced growth across segments so that more volatile businesses, such as trading and investment banking, do not account for an outsized portion of profits. We believe the recent share price undervalues the franchise, and we are maintaining our target price of \$30, which is 12.2-times our EPS estimate for 2021, when we believe an earnings recovery will be underway. BAC shares are down 19\% over the past year, versus a 7\% increase for the broad market. On July 15, the company reported 2Q20 earnings of \$0.37 per share, down from \$0.74 a year earlier but above the \$0.27 consensus. Net revenues were down 3.3\%, to \$22.3 billion, as lower net interest income from a narrower net interest margin was partly offset by greater investment banking and trading revenues. Net charge-offs were 0.45\% of average loans, up from 0.38\% the prior year. With \$1.1 billion of net charge-offs and a \$5.1 billion loan loss provision, there was a substantial reserve build as the company prepares for credit losses related to the coronavirus impact on the economy. First-half provisions totaled \$9.9 billion.      Noninterest expenses were \$13.4 billion, up 1\% from the prior year, and pretax income declined 58\% to \$3.8 billion, mostly reflecting the large loss provision.       Bank of America has four operating segments: Consumer Banking, Global Wealth and Investment Management, Global Banking, and Global Markets. Second-quarter results by segment follow below. In Consumer Banking, 2Q pretax income was down sharply to \$94 million, from \$4.4 billion a year earlier, hurt by a 19\% decline in revenues from a contraction in the net interest margin due to lower interest rates, and a three-fold increase in the credit loss provision in advance of expected defaults related to the impact of the coronavirus...
\end{tcolorbox}
\caption{A sample context as input to the model.}
\label{fig:BAC-report}
\end{figure*}

\begin{figure*}
\centering
\begin{tcolorbox}[colback=orange!5, colframe=orange, title= LLM output]
    
"<|start\_header\_id|>assistant<|end\_header\_id|> \\ Here is the analysis and investment rating in JSON format: \\{ "answer": "bullish",\\ "reason": "Although Bank of America Corp. (BAC) faces near-term challenges due to the impact of the coronavirus on its earnings, we believe that the company's long-term fundamentals remain solid. The recent decline in share price presents a buying opportunity, and we believe that the company's focus on 'responsible growth' will enable it to expand its loan portfolio without taking on excessive credit risk. Additionally, the dividend is secure, and capital levels are sound. With a target price of \$30, we believe that BAC shares are attractively valued."Note: The investment rating is bullish because the analyst believes that the company has a strong long-term fundamental profile, and the recent decline in share price presents a buying opportunity. The analyst is maintaining a target price of \$30, which is based on a multiple of 12.3-times the estimated 2021 earnings per share."}
\end{tcolorbox}
\caption{A sample LLM output given context with analyst rating}
\label{fig:BAC-LLM-output-with-rating}
\end{figure*}

\begin{figure*}
\centering
\begin{tcolorbox}[colback=teal!5, colframe=teal, title= Context]
    
The quarter included interest margin contraction as lower rates take hold, and a sharply higher credit loss provision as the bank prepares for delinquencies from the fallout of measures taken to contain the coronavirus. We look for continued pressure on revenues (from reduced loan volumes, year-over-year margin contraction and lower fee-based income) in the second half of 2020 to keep earnings below the recent run rate. Looking beyond near-term challenges, management continues to focus on what it terms 'responsible growth.' We believe this may be seen in the company's ability to expand its loan portfolio without taking on too much credit risk, and to maintain balanced growth across segments so that more volatile businesses, such as trading and investment banking, do not account for an outsized portion of profits. We believe the recent share price undervalues the franchise, and we are maintaining our target price of \$30, which is 12.2-times our EPS estimate for 2021, when we believe an earnings recovery will be underway. BAC shares are down 19\% over the past year, versus a 7\% increase for the broad market. On July 15, the company reported 2Q20 earnings of \$0.37 per share, down from \$0.74 a year earlier but above the \$0.27 consensus. Net revenues were down 3.3\%, to \$22.3 billion, as lower net interest income from a narrower net interest margin was partly offset by greater investment banking and trading revenues. Net charge-offs were 0.45\% of average loans, up from 0.38\% the prior year. With \$1.1 billion of net charge-offs and a \$5.1 billion loan loss provision, there was a substantial reserve build as the company prepares for credit losses related to the coronavirus impact on the economy. First-half provisions totaled \$9.9 billion. Noninterest expenses were \$13.4 billion, up 1\% from the prior year, and pretax income declined 58\% to \$3.8 billion, mostly reflecting the large loss provision. Bank of America has four operating segments: Consumer Banking, Global Wealth and Investment Management, Global Banking, and Global Markets. Second-quarter results by segment follow below. In Consumer Banking, 2Q pretax income was down sharply to \$94 million, from \$4.4 billion a year earlier, hurt by a 19\% decline in revenues from a contraction in the net interest margin due to lower interest rates, and a three-fold increase in the credit loss provision in advance of expected defaults related to the impact of the coronavirus...
\end{tcolorbox}
\caption{A sample context as input to the model.}
\label{fig:BAC-report-wo-rating}
\end{figure*}

\begin{figure*}
\centering
\begin{tcolorbox}[colback=orange!5, colframe=orange, title= LLM output]
    
"<|start\_header\_id|>assistant<|end\_header\_id|> Here is my analysis and investment rating: Answer: \\ Neutral \\ Reason: While Bank of America faces significant challenges in the near-term due to the impact of the coronavirus on the economy, its focus on \"responsible growth\" and its ability to manage its credit risk are positives. The company's recent share price undervalues the franchise, and its financial strength rating is Medium-High. However, the company's earnings are expected to be hurt by lower revenues and higher credit loss provisions in the near-term, and its dividend is expected to be lowered in 2020. While I am neutral on Bank of America, I do believe that its shares could recover once the impact of the coronavirus passes and the company's underlying earnings power improves. Additionally, the company's financial strength and liquidity provide a buffer against potential shocks. However, investors should be prepared for a near-term period of reduced earnings and dividends, which could impact the stock's performance. 

\end{tcolorbox}
\caption{A sample LLM output given context without analyst rating}
\label{fig:BAC-LLM-output-wo-rating}
\end{figure*}

\end{document}